\documentclass[letterpaper, 10 pt, conference]{ieeeconf}  
\usepackage[T1]{fontenc}
\usepackage{cite}
\usepackage{graphicx}

\IEEEoverridecommandlockouts                    

\usepackage{graphics} 
\usepackage{amsmath} 
\usepackage{amssymb}  

\title{\LARGE \bf
UniLegs: Universal Multi-Legged Robot Control through Morphology-Agnostic Policy Distillation
}

\author{Weijie Xi$^{\ast 1}$, Zhanxiang Cao$^{\ast 2}$, Chenlin Ming$^{2}$, Jianying Zheng$^{3\dag}$ and Guyue Zhou$^{4\dag}$
\thanks{*This work is supported by Wuxi Research Institute of Applied Technologies, Tsinghua University under Grant 20242001120 and 
the National Natural Science Foundation of China under Grants 62273017.}
\thanks{$^{1}$Weijie Xi is with the School of Internet of Things Engineering, Department of Automation, Jiangnan University,  Wuxi, 214122, China. Email: xiweijie@foxmail.com}%
\thanks{$^{2}$Zhanxiang Cao and Chenlin Ming are with Shanghai Jiao Tong University, Shanghai, P.R. China. Email: \{caozx1110, mcl2019011457\} @sjtu.edu.cn}%
\thanks{$^{3}$Jianying Zheng is with the School of Automation Science and
Electrical Engineering, Beihang University, Beijing, China. Email: zjying@buaa.edu.cn}%
\thanks{$^{4}$Guyue Zhou is with the Institute for AI Indus
try Research, Tsinghua University, Beijing 100084, China. Email: zhouguyue@air.tsinghua.edu.cn}%
\thanks{$\ast$ Equal Contributions.}%
\thanks{$\dag$ Corresponding author.}%
}

\begin{document}

\maketitle
\thispagestyle{empty}
\pagestyle{empty}

\begin{abstract}
Developing controllers that generalize across diverse robot morphologies remains a significant challenge in legged locomotion. Traditional approaches either create specialized controllers for each morphology or compromise performance for generality. This paper introduces a two-stage teacher-student framework that bridges this gap through policy distillation. First, we train specialized teacher policies optimized for individual morphologies, capturing the unique optimal control strategies for each robot design. Then, we distill this specialized expertise into a single Transformer-based student policy capable of controlling robots with varying leg configurations. Our experiments across five distinct legged morphologies demonstrate that our approach preserves morphology-specific optimal behaviors, with the Transformer architecture achieving 94.47\% of teacher performance on training morphologies and 72.64\% on unseen robot designs. Comparative analysis reveals that Transformer-based architectures consistently outperform MLP baselines by leveraging attention mechanisms to effectively model joint relationships across different kinematic structures. We validate our approach through successful deployment on a physical quadruped robot, demonstrating the practical viability of our morphology-agnostic control framework. This work presents a scalable solution for developing universal legged robot controllers that maintain near-optimal performance while generalizing across diverse morphologies.
\end{abstract}

\section{Introduction}

\subsection{Background}
With the rapid advancement of legged robotics, deep reinforcement learning has demonstrated remarkable capabilities in developing robust locomotion controllers~\cite{miki2022learning, margolis2023walk, kumar2022adapting}. These methods enable robots to walk, run, jump, and even perform complex parkour maneuvers on challenging terrains. However, a fundamental challenge persists in efficiently developing controllers that can generalize across diverse robot morphologies while maintaining optimal performance for each specific design~\cite{gupta2022metamorph, wang2018nervenet}. Traditional approaches that simultaneously train across multiple morphologies often compromise performance, as they struggle to discover the optimal policy for each specific configuration~\cite{kurin2020my, yu2023multi}. This limitation necessitates extensive retraining and hyperparameter tuning for each novel morphology~\cite{xu2023cross, salhotra2023learning}, creating a bottleneck in the deployment of legged robots across varied applications.

\subsection{Related Works}
To address the challenge of cross-morphology control, researchers have explored various architectural approaches. Graph Neural Networks (GNNs) were early efforts to model robot morphological structures~\cite{wang2018nervenet, huang2020one, sanchez2018graph}, representing robots as graphs with joints as nodes and physical connections as edges. However, these methods often struggle with generalization across significantly different morphologies, limiting their applicability in diverse scenarios.

Transformer architectures have subsequently been proposed to model interactions between joints~\cite{kurin2020my, gupta2022metamorph, yu2023multi}, employing attention mechanisms to enable global information aggregation. Despite these advantages, when trained simultaneously on multiple morphologies, Transformer-based methods often converge to sub-optimal policies that compromise performance to accommodate diverse configurations~\cite{chen2024trakdis, mertan2024towards}. This prevents these approaches from discovering the truly optimal control strategy for each specific morphology.

Hierarchical control frameworks~\cite{hejna2020hierarchically, sharma2019third} decompose tasks into morphology-agnostic high-level policies and morphology-specific low-level controllers, but require retraining low-level controllers for each new robot. Additional strategies like domain randomization~\cite{andrychowicz2020learning, rudin2022learning} and adaptive learning~\cite{kumar2021rma, niu2022trust} enhance policy robustness but primarily address single-morphology transfer.

Knowledge distillation has emerged as a promising direction for transferring expertise across domains and architectures~\cite{hinton2015distilling}. TraKDis~\cite{chen2024trakdis} introduces a Transformer-based knowledge distillation framework for visual reinforcement learning, while Mertan and Cheney~\cite{mertan2024towards} propose a multi-morphology controller that distills knowledge into a Transformer-based architecture. Foundation models~\cite{o2024open, reed2022generalist} and approaches like DiffuseLoco~\cite{huang2024diffuseloco} demonstrate potential for universal controllers with zero-shot transfer capabilities. Despite these advances, achieving both cross-morphology generalization and morphology-specific optimal performance in a single control policy remains a fundamental challenge in legged robotics.

\subsection{Contribution}
This paper introduces a novel two-stage teacher-student framework for morphology-agnostic robot control through policy distillation. The main contributions are summarized as follows:

\begin{enumerate}
    \item We propose a unified morphology-agnostic control framework that first discovers optimal specialized policies for each distinct robot morphology and then distills this specialized expertise into a single universal controller. Our two-stage process preserves the unique optimal control strategies for each morphology, with the Transformer architecture achieving 94.47\% of teacher performance across training morphologies, significantly outperforming traditional multi-morphology training approaches while enabling scalable integration of new robot designs.
    
    \item We implement both Transformer-based and MLP-based architectures to validate the effectiveness of policy distillation. To fully unlock the potential of our approach, we leverage the attention mechanism's ability to efficiently capture and transfer optimal control principles across diverse leg configurations. Our Transformer-based method achieves state-of-the-art performance, demonstrating its superiority in morphological generalization tasks.
    
    \item We validate the feasibility of our approach through successful real-world deployment on a physical quadruped robot. The hardware experiments demonstrate that our distilled policy maintains the essential locomotion capabilities across various terrains and tasks, confirming that the knowledge transfer from simulation to reality is effective. This real-world validation proves that our morphology-agnostic control framework is not only theoretically sound but also practically viable for real robot applications.
\end{enumerate}

\section{Problem Formulation}

\subsection{Partially Observable Markov Decision Process}
We formulate morphology-agnostic legged robot control as a Partially Observable Markov Decision Process (POMDP). For a given robot morphology \( M_i \), the POMDP is defined by the tuple \( (S_i, O_i, A_i, T_i, R_i, \gamma) \). The state space \( S_i \subset \mathbb{R}^{n_s^i} \) represents the full system state, encompassing the robot’s configuration, terrain characteristics, and environmental conditions. The observation space \( O_i \subset \mathbb{R}^{n_o^i} \) consists of proprioceptive information, including joint positions, velocities, and IMU readings, while excluding privileged information. The action space \( A_i \subset \mathbb{R}^{n_a^i} \) is defined by the set of joint targets, with its dimensionality determined by the actuated joints of \( M_i \). The transition probability function \( T_i: S_i \times A_i \times S_i \to [0,1] \) models the system dynamics, governing how actions influence state transitions. The reward function \( R_i: S_i \times A_i \to \mathbb{R} \) is designed to encourage stable and efficient locomotion. Finally, the discount factor \( \gamma \in [0,1) \) determines the relative importance of future rewards in policy optimization.  

A fundamental challenge stems from the dimensional heterogeneity of \( O_i \) and \( A_i \), which differ across morphologies due to variations in joint and sensor configurations. This discrepancy prevents the direct transfer of policies between different morphologies, as the observation and action spaces are inherently mismatched. For all given morphology \( M_i \), the universal policy \( \pi_{\theta}: O \to A \) is optimized to maximize the expected discounted return:
\begin{equation}
J(\pi_{\theta}) = \mathbb{E}_{\tau \sim \pi_{\theta}} \left[ \sum_{t=0}^{T} \gamma^t R(s_t, a_t) \right],
\end{equation}
where $\tau = (s_0, a_0, s_1, a_1, \ldots)$ represents a trajectory generated by $\pi_{\theta}$.

\subsection{Universal Control Objective}
The objective is to develop a universal policy \( \pi_\theta(a | o, m) \) that can effectively control any morphology \( M_i \in M \), attaining performance comparable to morphology-specific policies while generalizing to previously unseen morphologies. To this end, and to fully leverage privileged information available in simulation, we employ a two-stage optimization framework based on a teacher-student architecture\cite{lee2020learning}.

Correspondingly, the optimization of \( \pi_\theta \) is formulated as a two-stage learning process. In the first stage, for each morphology \( M_i \), the teacher policy is optimized to maximize the expected return:  
\begin{equation}
\phi_i^* = \arg\max_{\phi_i} J_i(\pi_{\phi_i}).
\end{equation}

In the second stage, the student policy is trained to imitate the teacher across different morphologies. The optimization objective is defined as:  
\begin{equation}
\theta^* = \arg\min_{\theta} \sum_{i=1}^{N} \mathbb{E}_{o \sim \mathcal{D}_i} \left[ \mathcal{L}(\pi_{\phi_i^*}(o), \pi_\theta(o, m_i)) \right],
\end{equation}
where \( \mathcal{D}_i \) is a dataset of observations collected from morphology \( M_i \), and \( \mathcal{L} \) is a loss function that quantifies the discrepancy between the teacher and student policies.

\section{Methodology}

\subsection{Two-Stage Teacher-Student Framework}

\subsubsection{Overview}
We propose a two-stage teacher-student framework to develop a universal controller for a wide range of legged robot morphologies. In the first stage, specialized teacher policies are individually trained for each morphology through reinforcement learning. In the second stage, these policies are distilled into a single, universal student policy. This separation not only facilitates efficient training but also ensures that the universal policy can effectively generalize across different morphologies.

\begin{figure}[t]
    \centering
    \includegraphics[width=1\linewidth]{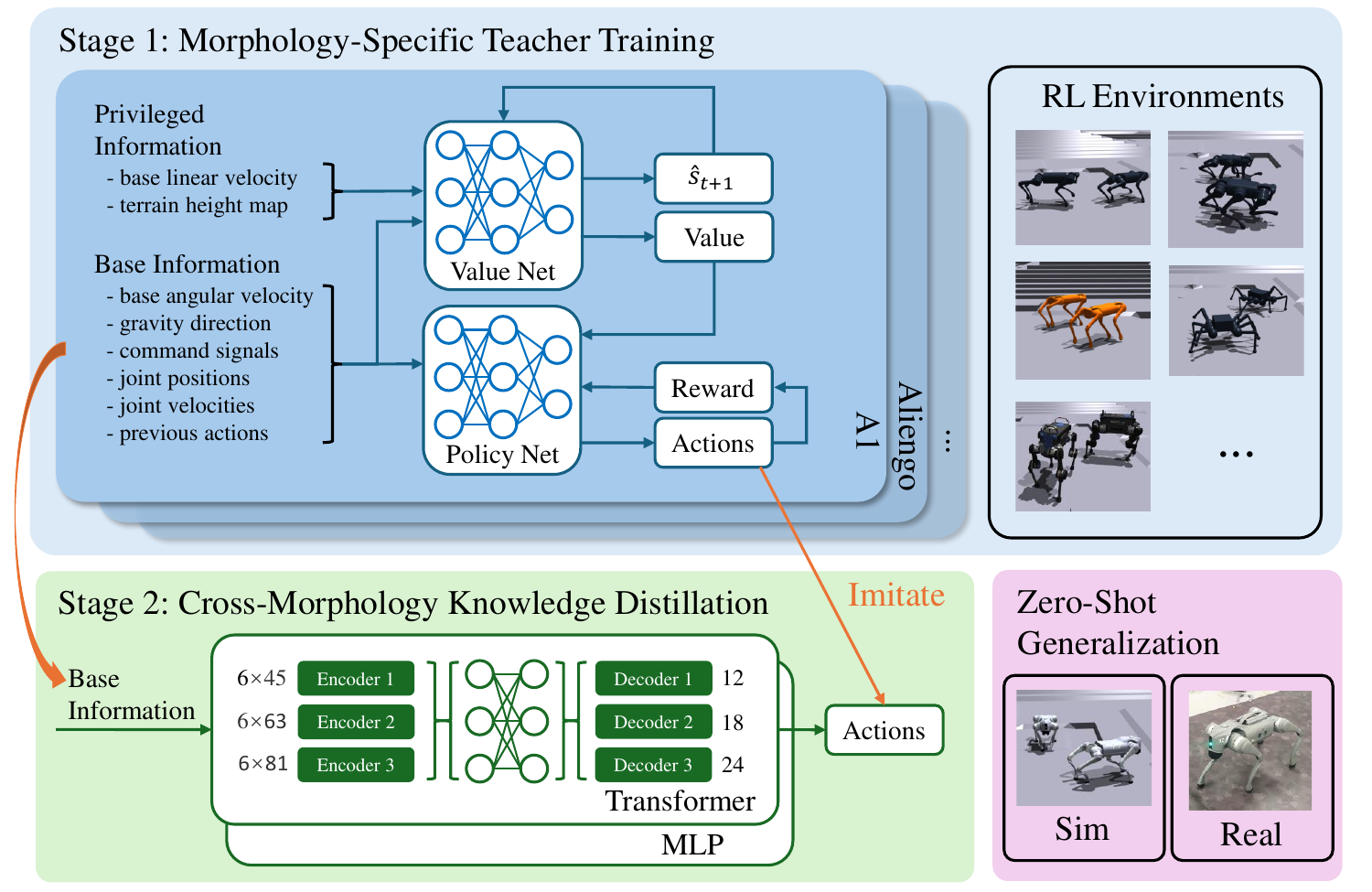}
    \caption{\textbf{Morphology-Agnostic Control Framework.} Our approach consists of two stages: (1) Morphology-specific teacher training where individual policies are learned for different robot morphologies (e.g., Aliengo, A1) using reinforcement learning with privileged information; (2) Cross-morphology knowledge distillation where a Transformer-based universal student policy learns to imitate all teacher policies using only base information available during deployment. The student architecture employs a series of encoder and decoder layers to process morphology-independent features, enabling zero-shot generalization to new robots in both simulation and real-world environments.}
    \label{fig:architecture}
\end{figure}

The teacher stage benefits from parallelization, sample efficiency, and performance optimization, as each teacher focuses exclusively on its specific morphology. Furthermore, reward functions can be tailored to individual morphologies, simplifying the design process. In the distillation stage, the universal policy extracts shared control principles, reducing overfitting and enabling few-shot or zero-shot generalization to unseen morphologies, as illustrated in Fig.~\ref{fig:architecture}.

\subsubsection{Stage 1: Morphology-Specific Teacher Training}
For each morphology $M_i$ in the training set $\mathcal{M}_{train} = \{M_1, M_2, \dots, M_N\}$, a teacher policy $\pi_{\phi_i}$ is trained using Proximal Policy Optimization (PPO) to maximize the expected discounted return:
\begin{equation}
\phi_i^* = \arg\max_{\phi_i} \mathbb{E}_{\tau \sim \pi_{\phi_i}}\left[\sum_{t=0}^{T} \gamma^t R_i(s_t, a_t)\right],
\end{equation}
where $\tau$ represents a trajectory generated by $\pi_{\phi_i}$. Each teacher policy is trained independently and in parallel, focusing on its specific morphology without requiring explicit morphology descriptors. Robustness is ensured by training under diverse environmental conditions, terrains, and disturbances.

\subsubsection{Stage 2: Cross-Morphology Knowledge Distillation}
The universal student policy $\pi_\theta$ is trained to mimic the behavior of all teacher policies through knowledge distillation:
\begin{equation}
\theta^* = \arg\min_{\theta} \sum_{i=1}^{N} \mathbb{E}_{o \sim \mathcal{D}_i} \left[ \mathcal{L}(\pi_{\phi_i^*}(o), \pi_\theta(o, m_i)) \right],
\end{equation}
where $\mathcal{D}_i$ is a dataset of observations from morphology $M_i$, and $\mathcal{L}$ measures the discrepancy between the teacher and student policies. This approach enables the universal policy to achieve morphology-agnostic control while preserving the performance of specialized teacher policies.

\subsection{Teacher Policy Training}

\subsubsection{Asymmetric Actor-Critic Framework}
We employ an asymmetric actor-critic architecture for each morphology $M_i \in \mathcal{M}$. The actor network $\pi_{\phi_i}(a_t|o_t)$ maps proprioceptive observations to actions, while the critic network $V_{\psi_i}(s_t)$ evaluates states using privileged information:
\begin{equation}
\pi_{\phi_i}: \mathcal{O}_i \rightarrow \mathcal{A}_i, \quad V_{\psi_i}: \mathcal{S}_i \rightarrow \mathbb{R}
\end{equation}
where $\phi_i$ and $\psi_i$ are the parameters of the actor and critic networks for morphology $M_i$, respectively.

\subsubsection{State and Observation Space}

As presented in Table~\ref{tab:obs_space}, the actor employs the robot's proprioceptive information as its observation, whereas the critic leverages additional privileged information available in simulation as its state to enhance the accuracy of value function estimation. In the table, \( n_i \) denotes the number of joints for morphology \( M_i \), and \( n_f \) represents the number of feet. To capture temporal dependencies, observations are further augmented with a history buffer of length \( H = 5 \).

\begin{table}[htbp]
\centering
\caption{Actor and Critic Input Features}
\label{tab:obs_space}
\renewcommand{\arraystretch}{1.2}
\begin{tabular}{lcc}
\hline
\textbf{Feature} & \textbf{Actor} & \textbf{Critic} \\
\hline
Base angular velocity $\boldsymbol{\omega}_b \in \mathbb{R}^3$ & \checkmark & \checkmark \\
Gravity direction $\mathbf{g} \in \mathbb{R}^3$ & \checkmark & \checkmark \\
Command signals $\mathbf{c}_t \in \mathbb{R}^3$ & \checkmark & \checkmark \\
Joint positions $\mathbf{q}_t \in \mathbb{R}^{n_i}$ & \checkmark & \checkmark \\
Joint velocities $\dot{\mathbf{q}}_t \in \mathbb{R}^{n_i}$ & \checkmark & \checkmark \\
Previous actions $\mathbf{a}_{t-1} \in \mathbb{R}^{n_i}$ & \checkmark & \checkmark \\
Base linear velocity $\mathbf{v}_b \in \mathbb{R}^3$ & $\times$ & \checkmark \\
Terrain height map $\mathbf{h}_{\text{terrain}} \in \mathbb{R}^{k \times k}$ & $\times$ & \checkmark \\
\hline
\end{tabular}
\end{table}

\subsubsection{Action Space}
The action space \( \mathcal{A}_i \) consists of target joint positions normalized to \( [-1, 1] \) and mapped to physical joint limits during execution:  
\begin{equation}
\mathbf{a}_t \in [-1, 1]^{n_i} \mapsto \mathbf{q}_{\text{target}} = \mathbf{q}_{\text{center}} + \mathbf{a}_t \cdot \mathbf{q}_{\text{range}}
\end{equation}  
where \( \mathbf{q}_{\text{center}} \) and \( \mathbf{q}_{\text{range}} \) represent the center positions and ranges for each joint. The target joint positions are then tracked by a PD controller with uniform gains set to \( K_p = 20 \) and \( K_d = 0.5 \).

\subsubsection{Reward Function}
Inspired by previous work\cite{luo2024moral}, the reward function used in our approach consists of a task-specific component, which measures the tracking error of linear and angular velocities, as well as several regularization terms. The details of the reward function are presented in Table~\ref{tab:reward_function}. To ensure a consistent evaluation across different morphologies \( M_i \), a unified reward structure and weighting scheme are applied to all robot configurations.

\begin{table}[htbp]
\centering
\caption{Reward function elements}
\label{tab:reward_function}
\renewcommand{\arraystretch}{1.2}
\begin{tabular}{p{2.6cm}p{3.1cm}l}
\hline
\textbf{Reward} & \textbf{Equation $(r_i)$} & \textbf{Weight $(w_i)$} \\
\hline
Lin. velocity tracking & $e^{-4(\mathbf{v}_{xy}^{\text{cmd}} - \mathbf{v}_{xy})^2}$ & 1.0 \\
Ang. velocity tracking & $e^{-4(\omega_{\text{yaw}}^{\text{cmd}} - \omega_{\text{yaw}})^2}$ & 0.5 \\
Linear velocity $(z)$ & $v_z^2$ & $-2.0$ \\
Angular velocity $(xy)$ & $\omega_{xy}^2$ & $-0.05$ \\
Orientation & $|\mathbf{g}|^2$ & $-0.2$ \\
Joint accelerations & $\ddot{\boldsymbol{\theta}}^2$ & $-2.5\times10^{-7}$ \\
Joint power & $|\boldsymbol{\tau}||\dot{\boldsymbol{\theta}}|$ & $-2\times10^{-5}$ \\
Body height & $(h^{\text{des}} - h)^2$ & $-1.0$ \\
Foot clearance & $(p_{f,z,k}^{\text{des}} - p_{f,z,k})^2 \cdot v_{f,xy,k}$ & $-0.01$ \\
Action rate & $(\mathbf{a}_t - \mathbf{a}_{t-1})^2$ & $-0.01$ \\
Smoothness & $(\mathbf{a}_t - 2\mathbf{a}_{t-1} + \mathbf{a}_{t-2})^2$ & $-0.01$ \\
Power distribution & $\text{var}(\boldsymbol{\tau} \cdot \dot{\boldsymbol{\theta}})^2$ & $-10^{-5}$ \\
\hline
\end{tabular}
\end{table}

\subsubsection{Training Algorithm}
We optimize the teacher policy using the Proximal Policy Optimization (PPO) algorithm. The actor and critic networks share the same hidden layer dimensions, set to \([1024, 768, 512, 256, 128]\), with layer normalization applied after each hidden layer to enhance training stability and performance.

\subsubsection{Domain Randomization}
To enhance robustness, we introduce randomness into various physical parameters, including mass, joint friction, and actuator strength, as well as terrain properties such as friction coefficients and compliance. Additionally, external disturbances, such as randomly applied forces, and sensor noise are incorporated into the training process. This strategy improves the generalization capability of the teacher policies, enabling them to adapt effectively to diverse real-world scenarios.

\subsection{Knowledge Distillation}

\subsubsection{Data Collection and Loss Function}  
For each morphology \( M_i \), a dataset \( \mathcal{D}_i \) is collected by executing the teacher policy \( \pi_{\phi_i} \) under diverse initial states, terrain conditions, and command variations. The universal policy is trained using a mean squared error (MSE) loss function, defined as:  
\begin{equation}
    \mathcal{L}_{MSE} = \frac{1}{N} \sum_{i=1}^{N} \mathbb{E}_{o \sim \mathcal{D}_i} \left[ \| \pi_{\phi_i}(o) - \pi_\theta(o, m_i) \|_2^2 \right].
\end{equation}

To ensure stable and efficient training, stochastic gradient descent is employed with learning rate scheduling, gradient clipping, and early stopping.  

\subsubsection{Student Model Architectures}
We investigate two distinct architectures for the student model: a multi-layer perceptron (MLP) baseline and a transformer-based architecture.

\textbf{Multi-layer perceptron (MLP)} consists of four fully-connected layers with dimensions [768, 1024, 1024, 512], employing ReLU activations between layers to capture non-linear relationships in the control mapping. This architecture directly maps the concatenated morphology descriptors and state observations to action outputs through sequential non-linear transformations, providing a straightforward baseline for morphology-agnostic control.

\textbf{Transformer-based architecture} maintains parameter parity with the MLP baseline while utilizing a different computational paradigm. It features a dimension of 256 for both input and output embeddings, with 8 parallel attention heads that process information in 32-dimensional key, query, and value spaces. The architecture comprises 3 transformer encoder layers, each incorporating multi-head self-attention mechanisms and feed-forward networks.

\section{Experimental Evaluation}

\subsection{Setup and Implementation}

\subsubsection{Robot Morphologies and Environment}
Our experiments evaluate morphology-agnostic control across diverse legged robot platforms. The training set comprises five distinct morphologies: three quadrupedal robots (Unitree A1, Aliengo, and ANYbotics ANYmal) with varying physical characteristics, plus six-legged and eight-legged variants to represent more complex kinematic structures. For zero-shot generalization testing, we employ the Unitree Go2 quadruped, which is excluded from the training process.

Simulations are conducted in NVIDIA Isaac Gym with a physics simulation frequency of 200 Hz and policy execution at 50 Hz. The environment features procedurally generated terrains including flat surfaces, slopes (5°–25°), steps (0.1–0.3× leg length), rough terrain, and gaps. We implement realistic actuator dynamics with position-controlled servos and apply a progressive difficulty curriculum to enhance policy robustness across challenging scenarios.

\subsubsection{Policy Architectures and Distillation Process}
We compare two student policy architectures: a Transformer-based model and an MLP baseline. Both architectures are designed with equivalent parameter counts to ensure fair comparison of their representational capabilities rather than capacity differences. The Transformer leverages self-attention mechanisms to dynamically model relationships between different joints and body parts, while the MLP employs a fixed connectivity pattern.

Knowledge distillation is performed using a comprehensive dataset containing 1 million state-action pairs per morphology, collected from expert teacher policies optimized specifically for each robot. The distillation process trains the student policies to reproduce the morphology-specific optimal control strategies across the diverse robot designs. Training is conducted over 500 epochs with a batch size of 2048, optimizing the student policies to capture the specialized locomotion strategies encoded in the teacher policies.

\subsubsection{Evaluation Metrics}
Performance is evaluated using normalized reward scores, where each morphology's specialized teacher policy defines the optimal performance baseline (1.0). This normalization enables direct comparison of transfer effectiveness across different morphologies despite their inherent differences in locomotion capabilities and control complexity. For the Unitree Go2 test morphology, we also train a specialized expert policy that serves as the normalization baseline. Importantly, this Go2 expert policy is used only for evaluation and is not included in the distillation process, providing a true measure of zero-shot generalization performance. This approach allows us to assess how effectively the distilled policies can transfer learned locomotion strategies to an entirely unseen morphology relative to its morphology-specific optimal performance.

\subsection{Results and Analysis}

\subsubsection{Teacher Policy Training}
\begin{figure}
    \centering
    \includegraphics[width=1.0\linewidth]{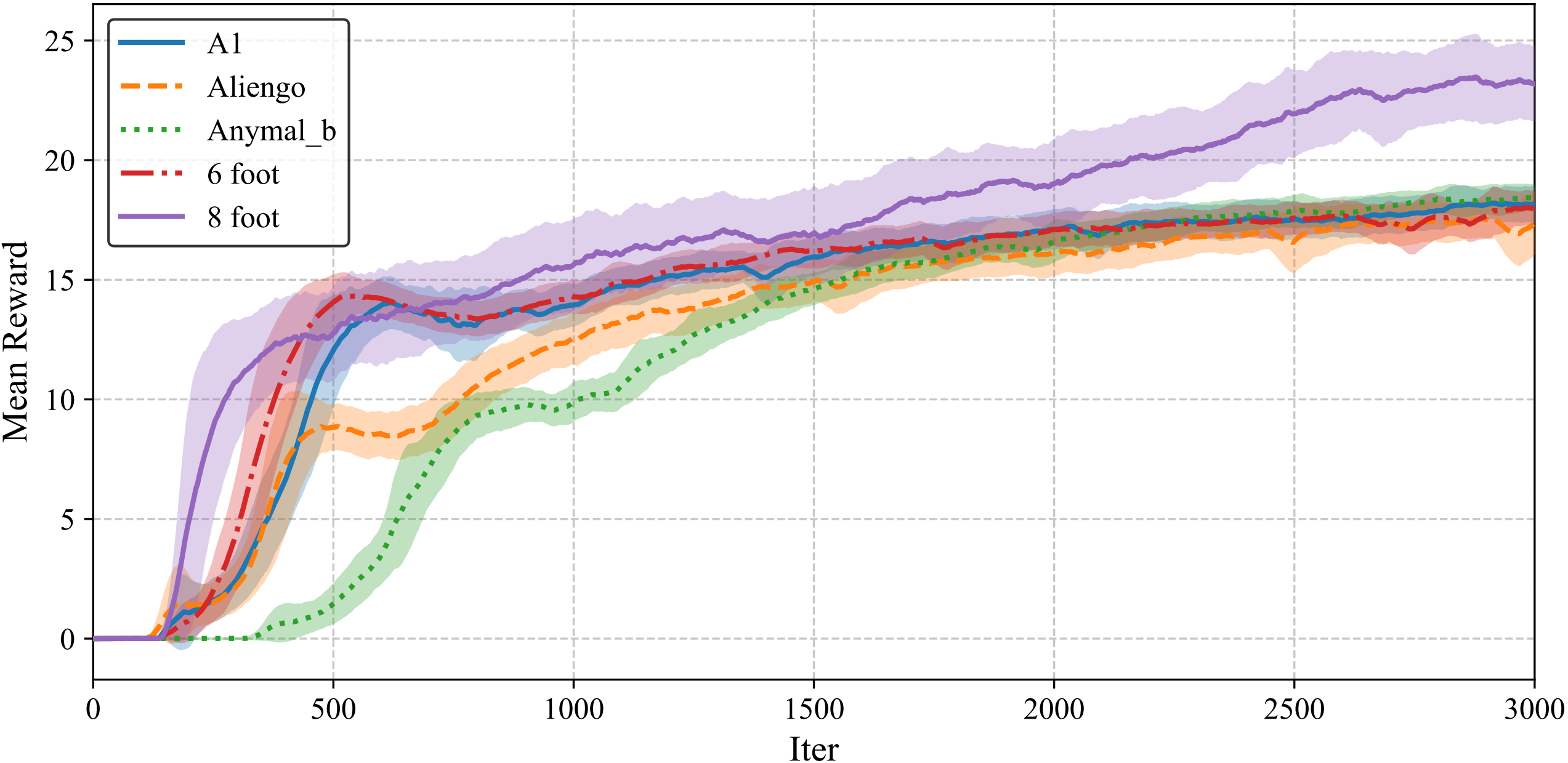}
    \caption{\textbf{Training curves of teacher policies across different robot morphologies.} The plot shows the mean reward as a function of training iterations for five distinct robot types: A1, Aliengo, Anymal\_b, 6-legged, and 8-legged robots. All morphologies demonstrate stable convergence after 3000 iterations, with the multi-legged robots (6-legged and 8-legged) achieving higher final rewards compared to quadrupedal robots.}
    \label{fig:train_reward}
\end{figure}

We first trained specialized teacher policies for each morphology using PPO. Fig.~\ref{fig:train_reward} shows the training curves of these teacher policies across different robot morphologies. All morphologies demonstrate stable convergence after 3000 iterations, with the multi-legged robots (6-legged and 8-legged) achieving higher final rewards compared to quadrupedal robots. This performance difference can be attributed to the inherent stability advantages of multi-legged platforms, which provide more contact points with the ground and redundant degrees of freedom for locomotion tasks.

\subsubsection{Performance Comparison}
Table~\ref{tab:performance_comparison} presents the normalized reward scores for teacher and student policies across diverse robot morphologies. The results demonstrate two key findings: the effectiveness of policy distillation in transferring morphology-specific optimal strategies, and the superior performance of the Transformer architecture compared to the MLP baseline.

\begin{table}[htbp]
\centering
\caption{Normalized Reward Scores for Teacher and Student Policies}
\label{tab:performance_comparison}
\renewcommand{\arraystretch}{1.25}
\begin{tabular}{lccc}
\hline
\textbf{Morphology} & \textbf{Teacher} & \textbf{STU (Transformer)} & \textbf{STU (MLP)} \\
\hline
A1 & 1.0000 & 0.8071 & 0.7465 \\
Aliengo & 1.0000 & 0.9703 & 0.9292 \\
Anymal & 1.0000 & 0.9683 & 0.8918 \\
6-legged & 1.0000 & 0.9936 & 0.9735 \\
8-legged & 1.0000 & 0.9843 & 0.9814 \\
\hline
\textit{Average (training)} & 1.0000 & 0.9447 & 0.9045 \\
\hline
Go2 (test) & 1.0000 & 0.7264 & 0.6929 \\
\hline
\end{tabular}
\end{table}

\subsubsection{Analysis of Distillation Effectiveness and Architectural Comparison}

Our experimental results demonstrate the efficacy of policy distillation in transferring morphology-specific optimal control strategies to generalized architectures. The distillation process aims to preserve the specialized optimal policies developed for each distinct robot morphology. Across training morphologies, the Transformer architecture consistently outperforms the MLP baseline, achieving 94.47\% versus 90.45\% of optimal teacher performance, representing a 4.02 percentage point advantage.

The effectiveness of transferring optimal strategies varies by morphology complexity. For the A1 quadruped, the Transformer and MLP capture 80.71\% and 74.65\% of the optimal control policy respectively. The Anymal with its FKBE leg configuration shows stronger transfer of optimal behavior at 96.83\% (Transformer) and 89.18\% (MLP), while Aliengo demonstrates 97.03\% and 92.92\% respectively. Multi-legged robots exhibit the most successful transfer of optimal strategies, with the Transformer achieving 99.36\% (6-legged) and 98.43\% (8-legged) of optimal teacher performance, and the MLP reaching 97.35\% and 98.14\% respectively. These near-perfect reproductions for complex morphologies indicate that the distillation process effectively captures the specialized optimal coordination patterns unique to multi-legged locomotion.

In zero-shot generalization to the unseen Go2 quadruped, performance decreases significantly, with the Transformer achieving 72.64\% and the MLP 69.29\% of the specialized optimal Go2 teacher performance. Despite this drop, the Transformer maintains a 3.35 percentage point advantage, demonstrating superior generalization capabilities in transferring optimal control principles even in zero-shot scenarios.

The Transformer's consistent advantage in preserving optimal strategies across all morphologies can be attributed to its self-attention mechanism, which dynamically models relationships between joints and body segments. While the MLP successfully captures many aspects of morphology-specific optimal strategies, attention layers provide greater flexibility in representing optimal control policies across diverse kinematic structures. This adaptability makes attention-based architectures particularly well-suited for transferring optimal control knowledge across varied robot designs.

The performance gap in zero-shot generalization to Go2 highlights the challenge of transferring learned optimal strategies to entirely new morphologies. Nevertheless, achieving over 70\% of specialized optimal teacher performance on an unseen robot demonstrates the potential of attention-based architectures for developing truly morphology-agnostic controllers that can approximate optimal control policies for legged locomotion.

\subsubsection{Real-world Deployment}

\begin{figure}[h]
    \centering
    \includegraphics[width=1\linewidth]{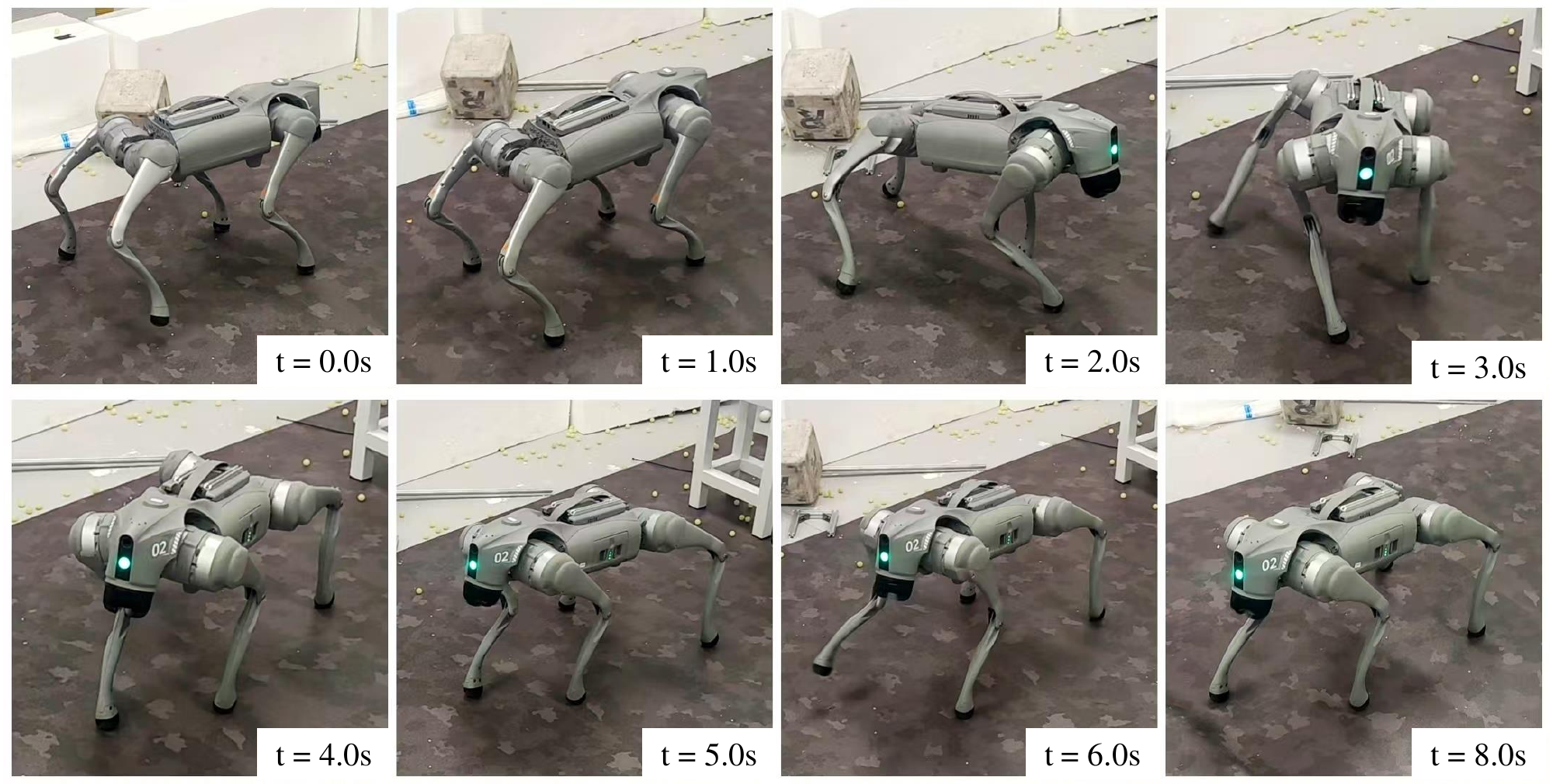}
    \caption{\textbf{Real-world deployment of the distilled policy on a Unitree Go2 quadruped robot.} The robot demonstrates stable locomotion capabilities in a laboratory environment, validating the sim-to-real transferability of our morphology-agnostic control approach.}
    \label{fig:real_deployment}
\end{figure}

To validate the practical applicability of our morphology-agnostic control framework, we deployed the distilled Transformer policy on a physical Unitree Go2 quadruped robot. As shown in Fig.~\ref{fig:real_deployment}, the robot successfully executed the learned locomotion skills in real-world environments. The deployment process involved transferring the trained policy weights to the onboard computer of the Go2 platform without any additional fine-tuning.

The physical experiments demonstrated notable consistency with our simulation results, with the robot exhibiting stable gait patterns and effective locomotion capabilities. The Go2 platform, which was not included in the training morphologies, serves as a critical test case for zero-shot transfer in real-world conditions. Despite achieving only 72.64\% of the specialized teacher performance in simulation, the deployed policy successfully bridged the reality gap, enabling functional locomotion behavior. This successful deployment confirms that our distilled policy can generalize not only across different morphologies in simulation but also transfer effectively to physical robots. The observed locomotion performance aligned with our simulation predictions, with the robot maintaining balance and responding to velocity commands reliably, further validating our knowledge distillation approach for universal legged locomotion control.

\section{Conclusion}
This paper presents a two-stage teacher-student framework for morphology-agnostic robot control through policy distillation. We first train specialized optimal controllers for distinct robot morphologies. Subsequently, we distill their expertise into a unified Transformer-based architecture. Our approach effectively bridges the gap between morphology-specific optimization and generalized control. Experimental results demonstrate that our distilled controller achieves 94.47\% of the optimal teacher performance across training morphologies and 72.64\% performance on entirely unseen robot designs.

The Transformer's attention mechanism models relationships between different components of robot morphologies with high efficacy. This capability enables flexible adaptation to varied kinematic structures. The successful reproduction of optimal control strategies across diverse robot designs confirms this advantage. Furthermore, real-world deployment validates the practical feasibility of our approach, demonstrating effective knowledge transfer from simulation to physical robots.

In conclusion, our two-stage distillation approach provides a practical pathway toward universal robot controllers that operate across diverse morphologies without compromising performance. The method preserves morphology-specific optimal control strategies while enabling generalization to new designs. This work highlights the potential of morphology-agnostic control policies for legged locomotion. It offers a promising solution for deploying legged robots across diverse applications without requiring bespoke control solutions for each morphological configuration.

\addtolength{\textheight}{-12cm}   

\bibliographystyle{IEEEtran}
\bibliography{IEEEabrv, references}

\end{document}